# Calculating Semantic Similarity between Academic Articles using Topic Event and Ontology

Ming Liu [a,1], Bo Lang [a] and Zepeng Gu [a]

[a] *State Key Laboratory of Software Development Environment, Beihang University, Beijing, China*

**Abstract.** Determining semantic similarity between academic documents is crucial to many tasks such as plagiarism detection, automatic technical survey and semantic search. Current studies mostly focus on semantic similarity between concepts, sentences and short text fragments. However, document-level semantic matching is still based on statistical information in surface level, neglecting article structures and global semantic meanings, which may cause the deviation in document understanding. In this paper, we focus on the document-level semantic similarity issue for academic literatures with a novel method. We represent academic articles with topic events that utilize multiple information profiles, such as research purposes, methodologies and domains to integrally describe the research work, and calculate the similarity between topic events based on the domain ontology to acquire the semantic similarity between articles. Experiments show that our approach achieves significant performance compared to state-of-the-art methods.

## 1. Introduction

Text semantics matching is widely used in many applications such as machine translation, automatic question answering, and knowledge retrieval. It also has great significance in plagiarism detection, automatic technical survey, citation recommendation and research trend analysis in academic domain. The issue of text semantics, such as word semantics and sentence semantics has received increasing attentions in recent years. However, rare research focuses on the document-level semantic matching due to its complexity. Long documents usually have sophisticated structure and massive information, which causes hardship to measure their semantic similarity, and there is even no public available dataset as far as we know.

Large text units are composed of small text units. The semantics of long document can be derived from the combination of small text unit semantics. Many recent studies follow this thought to acquire the semantic similarity between larger text units. For example, the sentence semantic similarity can be achieved from the integration of semantic similarities between word pairs from two sentences [1,2]. Besides lexical semantics, global sentence level features are also considered for acquiring the semantic similarity between sentence [1-8]. However, those studies only focus on short texts by lexical semantics and sentence-level features, which are still far from the capability of document-level semantic similarity.

---
[1] Corresponding Author, Ming Liu, State Key Laboratory of Software Development Environment, Beihang University, No.37 Xueyuan Rd., Haidian District, Beijing, China. E-mail: liuming@nlsde.buaa.edu.cn.

The study focusing on semantic similarity between documents is relatively rare. Existing methods of document-level similarity mainly focus on information retrieval in surface level rather than semantic level understanding. Those conventional similarity metrics [9-11] measure the document similarity by statistics or morphology of words, neglecting documents' structure and meanings of their words, such as vector space model (VSM) [12]. VSM regards each document as a collection of words and measures the document similarity mainly based on the presence of words, e.g., there are two snippets of text: "Jack borrowed a book from the teacher" and "The teacher borrowed a book from Jack". VSM regards the two texts as equal, but actually they have opposite meanings. Latent Dirichlet Allocation (LDA) [13] is proposed for the deep analysis of documents based on the difference of topic distributions over documents, which can be used to measure document-level semantics. There are also researches [14-19] trying to add external knowledge to the document representation, which enrich the content by adding the relevant terms from knowledge resources. However, these methods still suffer from problems like computational complexity and representational opaqueness.

Long documents contain many topic transitions and different focuses, which makes it difficult to capture their core semantics. However, we believe that those topics in a document are coherent, and those correlations can be obtained by the comprehensive analysis on various factors of a document. Hence, we represent the core semantic issue in each document as an event which is called a Topic Event (TE). TE is the structured summary extracted from each document, which contains the comprehensive key elements of the document.

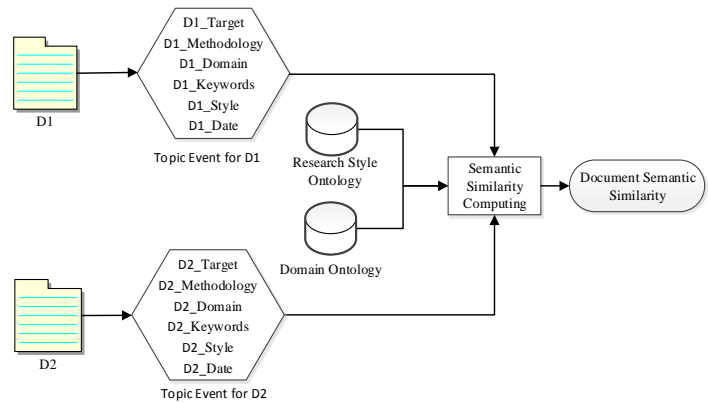

Fig. 1. TE-based document semantic matching

The core semantics of an academic article is the authors' research work. We construct TE based on the article structure and using multiple information fields such as research targets, methodologies, keywords and domains, which can integrally describe various facets of the research work. Hence, the semantic similarity between academic papers can be measured by TE similarity, which is shown in Fig. 1. For achieving high accuracy in TE similarity calculating, we also develop and utilize the research style ontology and domain ontology in the academic domain. In order to make our approach more practicable, we present the method of how to automatically construct a TE. To verify performances, we construct an evaluation corpora by manual annotation using ACL Anthology Network (AAN) corpora [20]. Experiments demonstrate that our methods gain outstanding performance, and the results are also more conforming to human's comprehensions. To summarize, the main contributions of our work are as follows:

—We propose the idea of constructing the topic events as the semantic representations of long documents, and give a general method for topic event similarity calculation.

—We develop and construct the research style ontology and the domain ontology for academic articles, and these two knowledge resources can facilitate the semantic extraction and similarity measurement procedures of the topic event effectively.

—We provide an ontology-based automatic TE construction method without labeled data in a specific domain, and utilize this method in the topic events construction of computational linguistic documents.

—We introduce a document semantic matching corpus with fine-grained annotations for the first time, which can serve as the ground truth for the evaluation of document-level semantic matching researches.

The remainder of the paper is organized as follows. In Section 2, we describe the related work. Section 3 provides the sketch of the topic event in academic domain. In Section 4, an automatic construction method of topic event is given. In Section 5, we describe the similarity computing method of the topic event. In section 6, the domain ontology and evaluation corpora in the computational linguistics domain are constructed, and then the experimental evaluations are given. Finally, we conclude the paper in Section 7.

**2. Related work**

The document-level semantic similarity research is a new arising area, which is thought can be enlightened by the concept and short text semantic matching methods. The concept-level semantics is the bricks of document semantic comprehension, and the short text-level similarity is the current most active area, which throws light on the document-level semantic similarity.

In general, concept-level semantic similarity can be measured by knowledge-based methods and corpus-based methods. Knowledge-based methods mainly utilize the path between concepts in knowledge resources to indicate their semantic similarity [21,22]. Lin [23] and Resnik [24] measured the semantic similarity by the information content ratio of the least common subsumer of the two concepts. Instead of directly exploiting the graph distance in knowledge resources, several researches [25,26] manufactured concept vectors according to a set of ontology properties for concept semantic similarity. Corpus-based methods assume that words with similar meaning often occur in similar contexts, Latent Semantic Analysis (LSA) [27] represents words as compact vectors via singular value decomposition (SVD) on the corpus matrix, and GloVe [28] reduced computational costs by training directly on the non-zero elements in corpus matrix. Turney [29] measured semantic similarity with the point-wise mutual information. Web can also be regarded as a large corpus, and the Google Distance [30] utilizes the number of words co-occur in web pages for concept semantic similarity. Mikolov [31] proposed the word embedding approach with neural network to capture the word semantics occurred in a fixed size window.

A challenge for determining semantic similarity of short texts is how to go from word-level semantics to short-text-level semantics. A direct approach is utilizing weighted sum of word-to-word semantic similarity, and [1,2] used the greedy word alignment method to form the sentence semantic similarity. Banea et al. [3] measured the semantic similarity between text snippets infused with opinion knowledge. D Ramage et al. [6] constructed a concept graph using words of each text snippet, then measured the similarity between two concept graphs. Recently, the SemEval conference released the Semantic Text Similarity (STS) task especially for the boom of short text semantics, and regression models are adopted by most teams [5,8] to predict the similarity score, and lexical feature and syntactic features are exploited. The paragraph vector [32], which is similar to word embedding, is also proposed with neural network to measure the semantic similarity between short texts.

The research directly related to document semantic similarity is rare and undeveloped. Traditional similarity researches such as the TF-IDF [12] converted documents to vectors via word counting, and measured the similarity between documents by the similarity of vectors. Academic articles can be regarded as semi-structured texts, which contain many structured annotations besides plain text. The similarity between academic articles can be measured with the help of annotated information. Martin et al. [33] fused the structured information, such as authors and keywords with traditional text-based measures for academic articles similarity. [9-11] regard articles and their citations as an information network. The similarity between articles turns to the similarity of two entities in information network. Unfortunately, the above conventional similarity methods aim at document indexing in surface level, rather than semantic level document understanding.

There are several studies trying to add external knowledge to achieve the semantic representation of documents. [14,15,17] enriched the content by adding relevant terms from knowledge resources, which aim at improving the quality of the document clustering results. [18,19] extracted relation triples from source document and added entity relations from background knowledge to construct triple graph for document enrichment. Schuhmacher and PonzeRo [16] proposed a graph-based semantic model for representing document content, which added knowledge to the document representation by linking the entities in the document to the DBpedia knowledge base. Those methods acquire fine-grained relationships among entities and generate a knowledge-rich document model, and the semantic similarity is calculated by the graph edit distance. However, above methods are lack of interpretation. Furthermore, there are rare entities such as persons, organizations and place names inside the content of academic articles, which makes those methods unsuited.

Long documents such as academic articles usually have several focuses and vast amount of words. LDA [13] obtains the semantic relatedness among different concepts via topics and regards each document as a distribution over a set of topics. Thus LDA can be used in the semantic analysis of long documents. Muhammad Rafi [17] defined a similarity measure based on topic maps in the document clustering task. The documents are transformed into topic maps based coded knowledge, and the similarity between a pair of documents is represented as a correlation between their common patterns. M. Zhang et al. [34] enriched document with hidden topics from external corpora, and measured the document similarity in text classification task with the similarity of topic distributions. Along the direction of topic model, [35,36] measured the semantic similarity between documents based on the divergence of topic distributions, which can be calculated by the Kullback-Leibler (KL) distance, and the LDA-based method is befitting in the semantic similarity task between academic articles.

## 3. Topic event

What can be used to convey the main semantics of a long document? The task is complicated and would not benefit directly from the simple accumulation of massive concept semantics. To get a global comprehension of a document, it's necessary to extract key information from large amount of words and form a core semantic of a document.

*3.1. Topic event structure*

In academic domain, articles are used to convey the research progress. Most academic articles have normative formats and regular structures, and the research work comprises similar profiles, which can be described in a uniform topic event structure.

The most notable points of an academic paper are their purposes, methods and results, which convey the main information of the research work. Keywords can fuzzily represent the core semantics and acquaint readers with the general research cognition. The domain of research issues indicates the research branch, and the type of research work reflects the research style and difficulty. The publication dates imply different stages of research issues. We define the above factors as the primary elements in a topic event, and the structure of the topic event is defined in Fig. 2.

The items annotated with * is crucial and essential, which cannot be vacant, while other items are optional. *Eid* and *Did* are essential identifications for topic event and the corresponding paper. *Style* indicates the research type, and the elements such as *Domain*, *Target*, *Methodology*, *keywords* are terminologies extracted from papers, while the elements of *Conclusion*, *Background*, *Performance* and *Forecast* are key sentences in papers.

```
Topic Event={
    *  Eid: essential ID for Topic Event;
    * Did: essential ID of the corresponding paper;
    * Target: essential purpose of the research;
    * Methodology: techniques used in the research;
    * Domain: essential category of the research issue;
    * Style: essential type of the research manner;
    *Keywords: keywords describing the article.;
    *Date: essential date of publication;
     Metadata:
        {author; organization; citation; publication; venue}
    Name: optional name of the developed system;
    Object: optional objects being researched;
    Tools: optional tools used in the research;
    Feature: distinguishing characters in research method;
    Conclusion: optional conclusion sentences;
    Background: optional supporting projects;
    Forecast: optional sentences for future research;
    Performance: the performance of the research;
    Dataset: optional datasets used in the research;
    }
```

Fig. 2. Structure of topic event

*3.2. Research style ontology*

The styles of research work implicate important semantics. They can reflect the variance of researches in difficulties, ways and types. For example, E1 and E2 are two research work as follows.

E1: The authors survey the techniques around a certain issue and summarize them as an academic article.

E2: The authors focus on a certain problem and propose a solution, the process and result of the solution are written as an academic paper.

Table 1 Details of the research style ontology

| Type | Remark | Example |
| --- | --- | --- |
| Theoretical Origination | Proposing original approaches | Latent Dirichlet Allocation |
| Methodology Improvement | Improving some methodologies or theories | Improving LDA Topic Models for Microblogs via Tweet Pooling and Automatic Labeling |
| System Implementation | Implementing some systems or tools | TEXTRUNNER:Open Information Extraction On the Web |
| Issue Solution | Solving some problems with existing methods | Biological Event Extraction using Subgraph Matching |
| Survey | Surveying certain research issues | An Overview of Event Extraction from Text |
| Analysis | Analyzing certain issues | a comparison of approaches to large-scale data analysis |
| Phenomenon Discovery | Uncovering certain conclusions | The Role of Research Leaders on the Evolution of Scientific Communities |

There are distinct differences between E1 and E2. E1 is a paper of Survey class, while E2 is a paper of Issue Solution class. In general, E2 has more innovation and difficulty than E1, and they have different values. E1 is suitable for beginners to get basic knowledge, while E2 is more suitable for inspiring experienced persons. Therefore, the type of an academic paper is an important factor in expressing its semantics. To express the knowledge implicated by types of research styles, we first develop the style categories of topic event, which is constructed in Fig. 3 by using protégé[37]. Each definite style of each research work is shown and explained in Table 1. The research style ontology implicates the relations between different styles of research work, which can be used to measure the semantics between different academic researches.

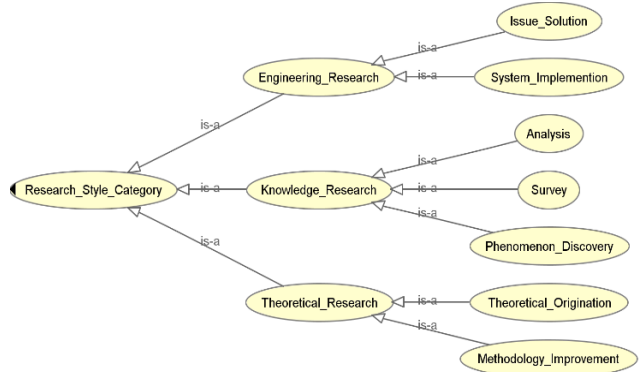

Fig. 3. Hierarchy of research style ontology

## 4. Automatic construction of topic event

*4.1. Overview*

As shown in Figure 2, we need to extract the *Target*, *Methodology*, *Domain*, *Style*, *Keywords* and *Date* from a document to construct the topic event. Present work on extracting structured representations of events has focused mainly on newswire texts, which utilize labeled entities, time expressions and values occurred in target sentences as candidate event items. The traditional event extraction problem is regarded as a classification problem with the help of labeled data. However, there is rare labeled data for training an extraction model, and almost no labeled entity can serve as the candidate terminologies in a specific domain.

Academic literatures have several characteristics which present unique challenges and opportunities for event item recognition. There are many structured annotations in academic articles such as the citations, authors, date of publication, keywords and journals, which are obvious to extract and can be used to enrich the topic event [38]. Notwithstanding, many important items of topic event are hidden in the unstructured content of academic articles. The main work of extraction is to identify terminologies such as *target*, *methodology*, *domain* and *style* in the article content. To solve the problem of lacking labeled event data, we propose an ontology and pattern-based extraction method.

Academic articles usually have clear topics and purposes, and there are many regular syntactical structures in academic articles which give clues to event extraction. The process of topic event construction is described in Fig. 4. Firstly, we divide the academic articles into different sections and select the most significant sections for topic event extraction. Then we conduct basic natural language processing (NLP) such as sentence splitting and part-of-speech (POS) tagging on each implicated sentence in the selected sections. Thirdly, we choose all the noun phrases (NP) in each sentence as candidate topic event items as well as a limited terminology list derived from the domain ontology. After that, the best event arguments are chosen from several candidates after pattern matching. At last, the extracted event items are delivered to the domain ontology to expand the related event semantic items. Generally, the domain ontology conveys the semantic relatedness among domain terminologies, which can provide the terminologies as well as their relationships in a specific domain. In this paper, the domain ontology can provide extern knowledge for the semantic comprehension of documents, and assist the procedure of topic event construction.

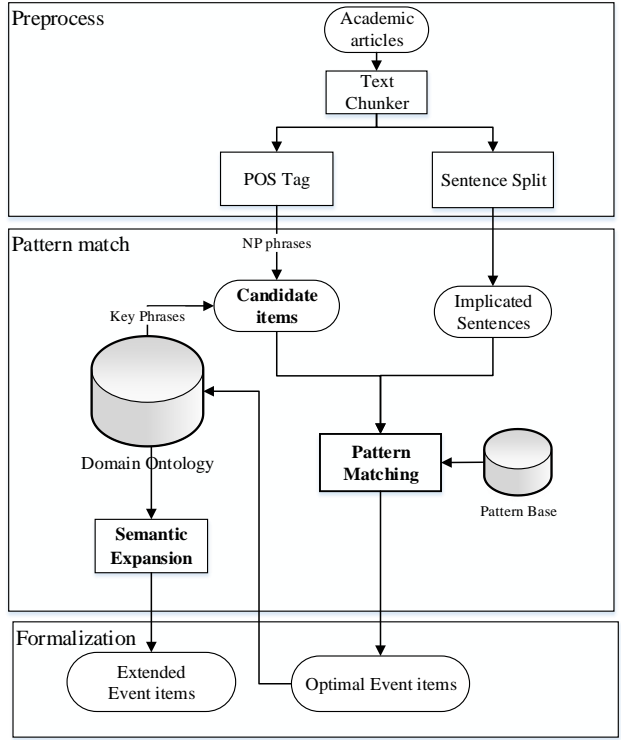

Fig. 4. Process of topic event extraction

*4.2. Event items recognition*

In detail, we divide each academic article into several chunks according to its outline, and the *Title*, *Abstract*, *Introduction* and *Conclusion* sections are believed to have a global description of the whole research work without much unnecessary detail. Then we identify the implicated sentences from above sections by trigger words. The implicated sentences may contain the event items, and we define a total of 95 trigger words for items extraction.

To capture the candidates of event items in implicated sentences, we leverage the domain ontology and some NLP processes. A terminology list derived from the domain ontology is used to find potential event items. However, another key issue is how to find a lot of unknown phrases in novel academic articles. To solve this problem, we conduct POS tagging on each sentence, and then utilize all the noun phrases as the candidates of event items to capture the unknown new phrases.

**Recognition of target and methodology** After acquiring the candidate terminology, the next step is to confirm which candidate is the best event item in each sentence. We develop the patterns for the extraction of *Target*, and *Methodology*. The extracting patterns are composed of pre-patterns and post-patterns, which are the patterns occurs frequently ahead of the event items and the patterns occurs frequently after the event items. Several typical extracting patterns of *Target* and *methodology* are shown in Table 2. There are altogether over 550 patterns for the extraction of *Target* and *Methodology*.

For example, the implicated sentence "*In this paper, we propose a supervised machine learning approach for relation extraction*" is identified by the trigger words "*propose*" from the introduction section of an academic article. It matches the preceding target pattern, i.e. "approach for". Thus, the terminology "*relation extraction*" is chosen as the target of this article.

Table 2 Typical extracting patterns for TE elements

| TE item | Patterns preceding the TE item | Patterns following the TE item |
|---|---|---|
| Target | the problem of<br>the task of<br>system to<br>survey on<br>approach for<br>framework for | overview<br>evaluation<br>track<br>system<br>called<br>process |
| Methodology | by use of<br>that using<br>which employ<br>takes advantage of<br>methods of<br>through an | based framework<br>method to<br>algorithm for<br>techniques to<br>is applied<br>performs much better |

Table 3 Typical patterns for research styles

| Research category | Pattern |
|---|---|
| Theoretical Origination | model<br>model for |
| Methodology Improvement | improving<br>improved<br>improvement |
| System Implementation | - , : ,corpus |
| Issue Solution | Using, based, focused, framework, expediting, hybrid, by, exploiting, incorporating, utilizing, methods, extracting, with, use, via, approach, measures |
| Survey | Overview<br>an overview of<br>survey<br>Introduction |
| Analysis | comparison<br>assessing<br>evaluation<br>evaluating<br>analysis<br>Challenges<br>the future for |
| Phenomenon Discovery | exploratory<br>exploratory of |

**Recognition of research style** We can see that articles of different research styles have different title characteristics from Table1. Many types of academic articles in computational linguistic domain can be diagnosed by the title of the paper. For example, each title of Issue Solution type is distinct, it will start with the abbreviated name of their software and connect the succeeding title with the punctuation ":" or "-", e.g., *"TEXTRUNNER : Open Information Extraction On the Web"*, *"URES : An Unsupervised Web Relation Extraction System"*. Most titles have featured words to distinguish their research styles, and we develop patterns to identify the topic research style in Table 3.

*4.3. Ontology-based semantic expansion*

Many items in a topic event are closely correlated such as the research purpose and domain, the adopted methodology and toolkit, the research object and dataset. Generally, the research purposes are the core issues of the academic articles and corresponding topic events, and the domain that an academic article belongs to is decided by its research purpose. When we extract the target of an academic article, we use the domain ontology to induct which domain it belongs to.

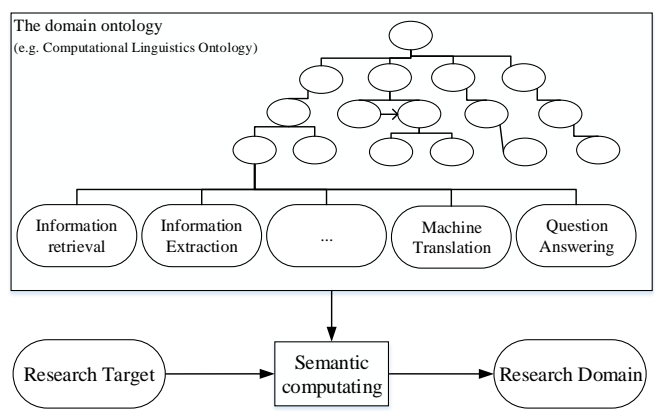

Fig. 5. Architecture of ontology-based semantic expansion

After extracting the research target, the semantic similarities between the target and each predefined domain concept are calculated based on the domain ontology, as shown in Fig. 5 which uses the Computational Linguistics Ontology as an example of domain ontology. The domain concept which has maximal semantic similarity with the target concept is chosen as the domain of the corresponding research article.

**5. Topic event similarity calculation**

*5.1. General framework*

Since topic event can represent the semantics of a document, the semantic similarity between documents can be achieved via the similarity between topic events. In this paper, we get the information hidden in each document via the topic event, and acquire the internal relevance among concepts from the domain ontology.

In our method, to compute the core semantics of papers in brief, six primary elements in topic event are employed according to their characteristics, which are *Target*, *Domain*, *Style*, *Methodology*, *Keywords* and *Date*. We use topic event ontology to measure the internal event similarity between different types, and the domain ontology is used to measure the internal semantic similarity between terminologies. We measure topic event similarity by the weighted sum of element similarities in corresponding event structure, and it also can be extended by metadata and other elements for a more detailed topic event similarity. Similarity between topic events E1 and E2 is defined in equation (1):

$$SimTEs(E_1, E_2) = \sum_{i=1}^{6} W_i \times S_i(L_{1i}, L_{2i}) \tag{1}$$

where $w_i$ is the weight of the *i*th element in the topic event, $S_i$ is the similarity function between the *i*th elements of $L_1$ and $L_2$. *L1* and *L2* are two topic events whose elements are defined as *L= {Target, Domain, Style, Methodology, Keywords, Date}*.

The topic events extract the core semantics of each academic articles. However, the meaning of event items could not be measured by their literal appearance. Background knowledge such as lexical meaning is necessary to comprehend the semantics of articles. To get the internal semantic relatedness among different terminologies, the terminologies in topic events should be linked to the concept nodes in knowledge bases to obtain their semantic relatedness. In the following section, we will introduce the concept linking method and the measurement of item semantic similarity.

*5.2. Domain ontology concept linking*

To calculate the internal similarity between terminologies, an important issue is to link the extracted terminologies to proper positions in the knowledge base. There are many synonyms in academic articles, and many concepts may be described by different terminologies in different papers such as "*cross linguistic retrieval*" and "*multilingual information retrieval*", "*text understanding*" and "*message understanding*", "*named entity recognition*" and "*named entity tagger*". When we construct the domain ontology, we labeled all the known synonyms of each concept in the node to facilitate entity linking.

Many automatically extracted terminologies wear trivial suffix and prefix, which may hinder the entity linking. To solve this problem, we use the edit distance to measure the string similarity and recognize the variations of the same concept. Since our ontology are constructed for the computational linguistics domain, and all of the concepts nodes are extracted from domain corpus, most terminologies will find the extract concept position in the domain ontology. We first derive a terminology list to form the domain ontology. When a terminology extracted from the academic articles comes, its edit distance with each of the terminology are calculated. The concept node with minimum edit distance will be regarded as the terminology node.

*5.3. Item similarity of topic event*

The items such as *Target*, *Methodology* and *Domain* are terminologies, and their semantic similarities can be measured by the domain ontology. The semantic similarity between Style can be measured via the research style ontology. The Date similarity can be measured by their interval. In brief, the similarity of those items in topic events can be measured by the following methods.

**Research Style Similarity** To measure the difference between different types of research work, the research style ontology described in Table 1 can be used. The *Style* similarity between different types of topic events is measured similar to Wu and Palmer method [22] based on the research style ontology shown in Fig. 3, and the formula is as equation (2):

$$Sim_{Etype} = \frac{2 \times depth(LCS)}{depth(Style_1) + depth(Style_2)} \quad (2)$$

where $Style_1$ and $Style_2$ means the types of two topic events. LCS is the least common subsumer of two style nodes.

**Terminology Similarity** Concept semantic similarity can be measured by knowledgebase. We evaluated several knowledge-based methods and found Wu and Plamer [22] method is suitable for the concept similarity in this domain. The contents of *target*, *Domain*, *Methodology* and *Keywords* are collections of terminologies, which can be measured by Wu and Palmer method based on the domain ontology or word embedding-based methods. The ontology-based concept semantic similarity is measured by equation (3):

$$Sim_{ec} = \frac{2 \times depth(LCS)}{depth(ec_1) + depth(ec_2)} \quad (3)$$

where $ec_1$ and $ec_2$ represent terminologies in the topic event.

When the corpus-based concept similarity method is used for terminology semantic similarity, the cosine similarity between terminology vectors can be used. The cosine similarity is defined in equation (4):

$$Sim_{ec} = \frac{TermVec_1 \cdot TermVec_2}{|TermVec_1||TermVec_2|} \quad (4)$$

**Date Similarity** Research issues keep evolving along time, and researchers will focus on different scientific issues in each period. We assume that the academic articles that have close date will be more similar, and the academic articles published far from each other will have less common points. Thus, date similarity can be measured by the time interval. *Years* and *months* are used to compute the similarity between two dates. We define the Date Similarity formula is as equation (5) .

$$Sim_{Date} = \frac{1}{1 + |(year_1 + \frac{month_1}{12}) - (year_2 + \frac{month_2}{12})|} \quad (5)$$

## 6. Experimental evaluation

### 6.1. Corpora construction

There are several public datasets used to evaluate the semantic similarity of short texts and sentences, such as MSPR [3], Michael D.LEE 50 corpus [7] and SEMILAR corpus [5]. However, no text of those datasets is more than 200 words, which could not validate the document-level semantic similarity. Hence, we construct the semantic similarity dataset between documents using the academic papers in computational linguistics domain. A set of paper pairs are generated from AAN corpus [20]. The paper pairs are annotated by both a 2-level and by a 5-level annotation as ground truth. Each paper pair is marked as 1 if semantically similar or 0 if dissimilar in 2-level annotation. In 5-level annotation a paper pair is marked by integers ranging from 1 to 5 according to their semantic similarity degree. If they are totally equal in semantic, similarity between papers will be annotated as 5, and if they have nothing to do with each other, similarity is annotated as 1. Twelve experts from our lab annotated and cross validated the coherence of 1021 pairs of documents. Each paper pair get a second annotation by a different person after the first annotation. If the second annotation is in accordance with the first one, it will be annotated as the ground truth, else a third person will annotate the paper pair to get the ground truth. In the end we got an annotated corpus with 1021 paper pairs. Now the corpus is public, and the url is: https://github.com/buaaliuming/DSAP-document-semantics-for-academic-papers/tree/buaaliuming-annotation .

*6.2. Computational linguistics ontology*

Common knowledge resources such as WordNet couldn't cover domain terminologies. In order to compute semantic similarity between terminologies, we manually construct the domain ontology to convey the semantics among different terminologies.

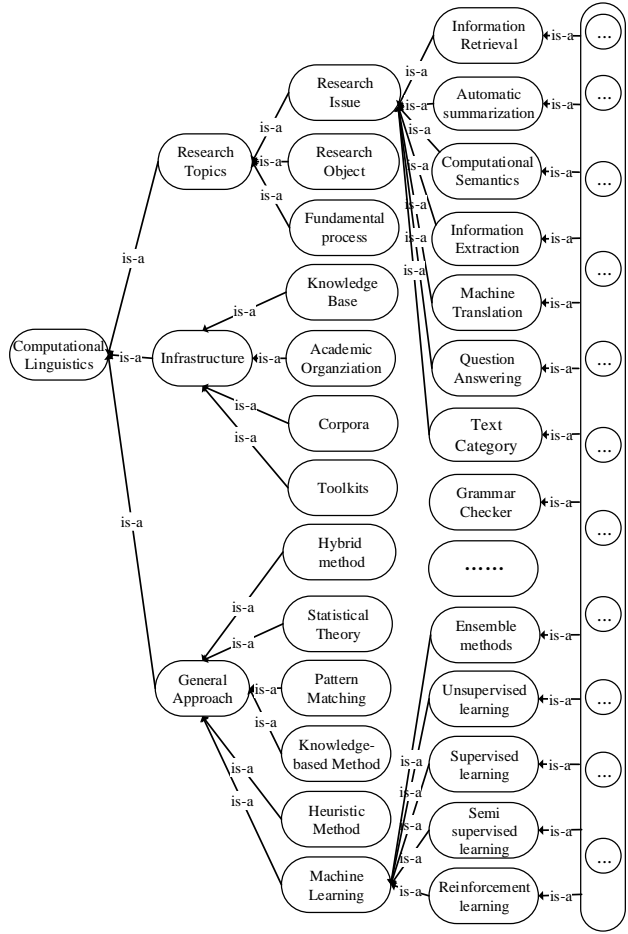

Fig. 6. Main structure of computational linguistic ontology

The Concepts extracted from AAN corpus [20] are used to manually construct a computational linguistics ontology. Currently our ontology includes 1195 concept nodes with a hierarchy of 9 depths, which will be expanded persistently in future. The Computational Linguistics Ontology (CL Ontology) architecture is shown in Fig. 6. The main relationships between concepts in the ontology are hyponymy. Synonyms are considered and annotated in the ontology concept nodes during the construction.

CL ontology is used to measure the similarity between computational linguistic domain concepts. According to the characteristic of computational linguistic domain, we design the ontology as three parts, they are: *Research Topic*, *Infrastructure* and *General Approaches*, and each part is enriched by more detailed descendant nodes. The node *General Approach* includes common methodologies used in computational linguistics, such as *machine learning*, *pattern matching* and *knowledge engineering*, etc. The node *Research Topic* includes *fundamental language processes*, *research issues* and *research objects*. *Fundamental language process* includes natural language processing, such as *word segmentation*, *syntactic parsing*, *POS tagging*, *lemmatization*, etc. *Research issues* include research hotspots such as *machine translation*, *text category*, *information extraction*, *information retrieval* and *speech recognition*, etc. The node *Infrastructures* contains the general toolkits, knowledge bases, corpora, and organizations in the field of computational linguistics.

*6.3. Experimental setup*

The experiments are conducted on the DELL OptiPlex390 machine, which has 8G memory and an I5-2400 CPU. Besides automatic constructed topic events, we also manually annotate the corresponding topic events of academic papers for experimental contrast. The LDA-based method is chosen as the baseline method. When computing the semantic similarity between terminologies in topic events, the LSA method is used as well as the ontology-based method. In brief, we conduct the following methods.

**LDA_2013** The LDA-based method [36] in 2013 is the most related research and is chosen as the baseline method for contrast. When performing the LDA-based method, several LDA models with different parameters are trained based on the AAN corpus. In the following presented results, we choose the LDA model with 200 topics for contrast, which achieves the best performance among different LDA models under the same conditions.

**TE_Onto** the TE semantic similarity method are conducted on the golden topic event annotations, and the concept semantic similarities are computed by our CL ontology.

**AutoTE_Onto** the TE semantic similarity method are conducted on the automatic extracted topic events, and the concept semantic similarities are also computed by our CL ontology. The AutoTE_Onto method is used as the comparison of TE_Onto method to measure the influence of automatic topic event extraction.

**TE_LSA** the TE semantic similarity method are conducted on the golden topic event annotations, and the concept semantic similarities are computed by the word vectors produced by LSA. The LSA method derives each word representation by SVD operation, and the semantic similarity between terminologies is captured by the similarity of the common topics. When the LSA method is used in the calculation of TE item, we constructed term-by-document matrix over the whole annotated corpus. The TE_LSA method is used as the comparison of the TE_Onto method to measure the effect of the Computational Linguistics Ontology.

**AutoTE_LSA** the TE semantic similarity method are conducted on the automatic extracted topic events, and the concept semantic similarities are also computed by the word vectors produced by LSA.

**TE item weights** The primary items of the topic event are *Target*, *Domain*, *Methodology*, *Style*, *Keywords* and *Date*, which are the essential items in Fig. 2. According to our experience, the research target is the crucial issue in each academic article. The research domains, types of research work and adopted methods in academic articles are important aspects, which are discriminative characteristics of each research work. While the keywords are a set of fuzzy description lack of definite semantics. The dates of publication imply the different ages of technology, which are less discriminative and not directly related to various research work. We set the weight of aforementioned items according to their importance, and in our experiments, weights of the items, i.e. *Target*, *Domain*, *Style*, *Methodology*, *Keywords* and *Date* are set to 0.3, 0.25, 0.25, 0.1, 0.05 and 0.05 respectively.

**Evaluation metrics** We choose the Pearson's correlation to measure the quality of semantic similarity scores. The larger Pearson's correlation is, the more correlated the predicted scores and the ground truth are. The Pearson's correlation is shown in equation (6).

$$\rho_{X,Y} = \frac{cov(X,Y)}{\sigma_X \sigma_Y} = \frac{E((X-\mu_X)(Y-\mu_Y))}{\sigma_X \sigma_Y} \tag{6}$$

In this work, *X* is the predicted semantic similarity score, and *Y* indicates the annotated semantic similarity value. The $cov(X,Y)$ represents the covariance of *X* and *Y*. The $\mu_X$ and $\mu_Y$ represent mean values of the variables *X* and *Y*; $\sigma_X$ and $\sigma_Y$ are standard deviations of *X* and *Y* respectively.

Since our corpus has both binary annotations and five-level annotations, we further set different thresholds to predict whether two documents are semantic similar. Both the Accuracy and F1-score can be the overall evaluation metric beside correlation. The Accuracy is shown it equation (7).

$$Accuracy = \frac{(TP+TN)}{(TP+TN+FP+FN)} \tag{7}$$

In equation (7), *TP* means the number of document pairs predicted to be similar that actually are similar document pairs. *TN* means the number of document pairs predicted to be dissimilar that actually are dissimilar. *FP* is the number of document pairs predicted to be similar that are actually dissimilar *FN* is the number of document pairs predicted to be dissimilar that are actually similar.

Accuracy means the general predict ability of a method, and F-score means the comprehensive performance of precision and recall. The F-score is shown it equation (8).

$$F\ score = \frac{(1+\beta)P}{(\beta P+R)} \tag{8}$$

The P means Precision and the R means Recall. In general, the value of β is 1, and the F score is annotated as F1-score in the following section.

*6.4. Results and discussions*

*6.4.1. Pearson correlation*

To verify the quality of different methods, we calculate the Pearson's correlations of 1021 similarity scores with the human annotated ground truth.

**Comparison with the baseline** The results in Table 4 show that our ontology-based TE method has distinct advantage on the baseline method, i.e. LDA_2013. The 5-level annotation has more detailed similarity levels; and the correlation scores with it are more convincing than the 2-level correlation scores. Our TE_Onto method achieves 4.1% (relative) improvement over the baseline method; when the topic events are extracted automatically, our AutoTE_Onto method can gain 5.8% (relative) improvement over the baseline method.

Table 4 Pearson's correlations with ground truth

| Correlations | 5-level Annotation Correlation | 2-level Annotation Correlation |
|---|---|---|
| TE_Onto | 0.559 | 0.461 |
| AutoTE_Onto | 0.568 | 0.456 |
| TE_LSA | 0.480 | 0.346 |
| AutoTE_LSA | 0.463 | 0.327 |
| LDA_2013 | 0.537 | 0.250 |

**The impact of ontology** Our TE_Onto method performs much better than the LSA-based TE method, and the AutoTE_Onto even shows 22.7% advantage over the AutoTE_LSA method when topic events are extracted automatically.

The LSA-based method measures the semantic similarity of concepts by common word topics or word co-occurrence, while the ontology-based methods measure the concept semantic similarity via accurate knowledge in ontology. Above results show that the knowledge resource such as the domain ontology are of great importance in the measurement of document semantics.

**The impact of TE extraction** The correlation scores of our methods on automatic constructed topic events and human annotated topic events are close. The performance of our methods with automatic extracted topic events is comparable to the golden topic events annotated by experts. The TE_LSA method only shows minor advantage on the AutoTE_LSA method, and the AutoTE_Onto method even performs little better than the TE_Onto method in 5-level annotation correlation. This shows that, the pattern-based extracting method can extract necessary information with proper precision in a specific domain, and the process of automatic extraction produces comparable performance to human annotation.

*6.4.2. Accuracy and F1-score*

Accuracy and F1-score can be the general evaluation metrics beside Correlation. The Accuracy indicates the general capability of a method to predict the right result; the F1-score is a balance of precision and recall, which indicates the comprehensive capacity of the method. In practical applications, different thresholds can be set to predict whether two documents are semantic matching. Generally, the best performance among different thresholds is considered as an important factor in evaluation. In the following experiments, different thresholds are set. Paper pairs are thought semantically similar if their similar score is greater than the given thresholds.

**Comparison with the baseline** As shown in Fig. 7 and Fig. 8, the Accuracy of our TE methods always has distinct advantage over the LDA_based method at different thresholds.

Our best F1-score is 0.639, while the best F1-score of the baseline method is 0.536. The F1-score of our AutoTE_Onto method outperforms the LDA_2013 at most thresholds, and the F1-scores of our TE_Onto methods with golden topic event show advantage over the baseline method when the threshold is less than 0.75.

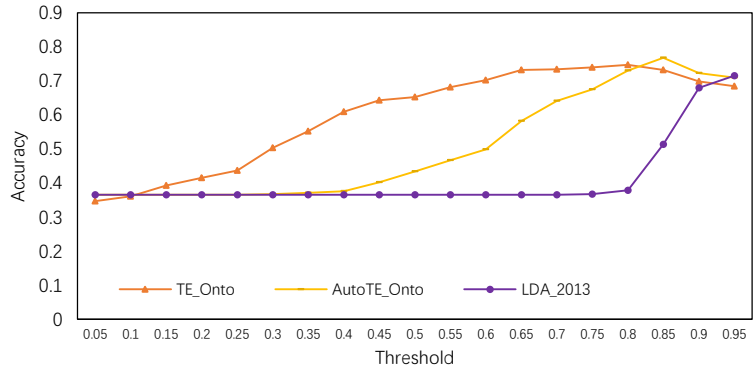

Fig. 7. Comparison of accuracy of our methods and the baseline method

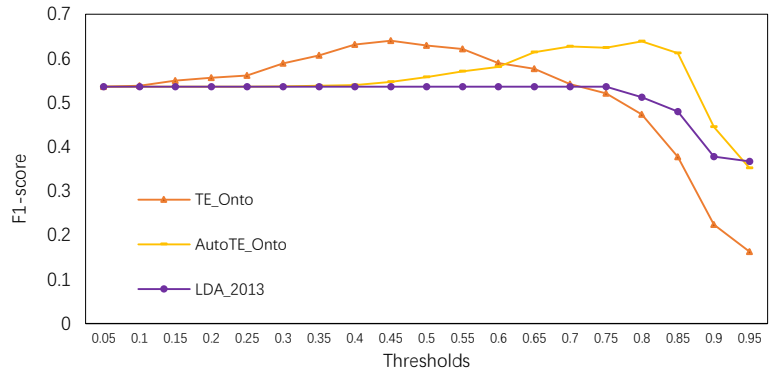

Fig. 8. Comparison of F1 of our methods and the baseline method

**The impact of ontology** We compare the performance of LSA-based TE method and the Ontology-based TE method. As shown in Fig. 9 and Fig. 10, the best Accuracy of LSA-based TE method is 0.712, while the best Accuracy of Ontology-based TE method is 0.768. The Ontology-based TE methods perform much better than the LSA-based TE methods when thresholds are more than 0.250. To sum up, the ontology-based TE methods perform better than LSA-based TE methods, which is in accordance with the result of the Pearson Correlation in the last subsection. Those results show that the ontology plays a crucial role in the document semantic similarity.

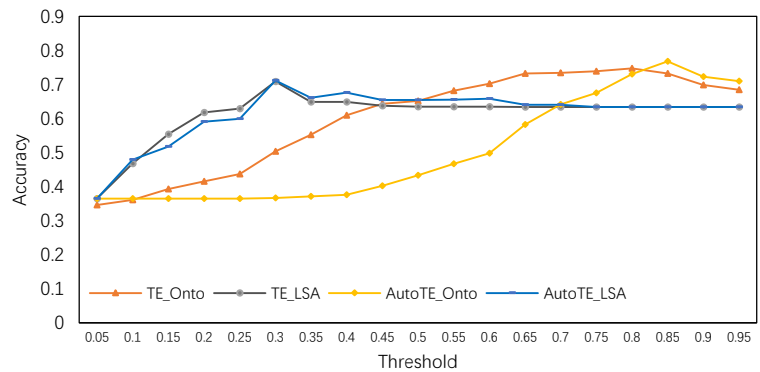

Fig. 9. Comparison of accuracy of Ontology-based TE methods and LSA-based TE methods

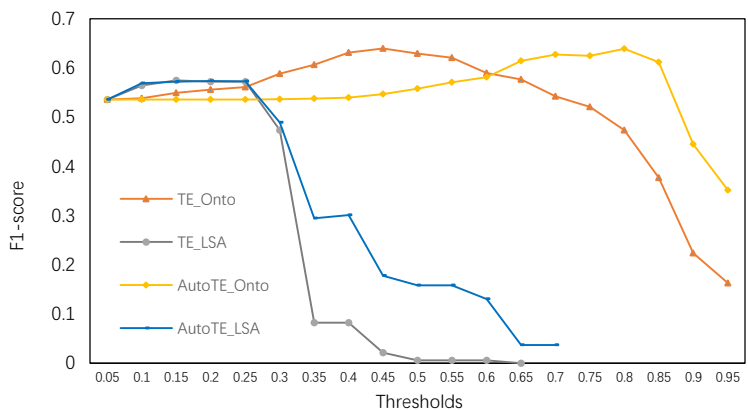

Fig. 10. Comparison of F1 of Ontology-based TE methods and LSA-based TE methods

**Discussion** Above experiments demonstrate that TE methods can yield much better performance in general. The similarity scores of TE_LSA method and AutoTE_LSA method range from 0.0 to 0.70, and there is no true positive result when the thresholds are above 0.70. Those low level similarity scores make the F1-scores of LSA-based TE method incalculable. Academic articles usually have long contents containing lots of duplicate terminologies, and the frequent terminologies in a specific domain trend to occur in many domain articles. Besides, academic papers tend to review related work. The LDA-based method determines the topic similarity by the word overlap. Hence, the similarity scores of LDA_2013 ranges relatively high from 0.7 to 1.0, which makes it perform good in recall but poor in precision, and depress its overall performance. Our methods hold the core semantics of a document directly via structured topic event, the similarity scores of TE methods fluctuate from 0.0 to 1.0 according the similarity of document pairs, and they are more discriminative and have better overall performance.

*6.4.3. Time and memory cost*

Each method in our experiment needs an off-line process. It's hard to measure and compare the off-line process of those costs in different environments and procedures. In this section we measure the running time when computing the document-level semantic similarity under the same condition. The average consuming time of TE methods is 0.002s and the memory occupation is about 100M, while the LDA_2013 costs 4.83s and occupies more than 8G memories. It's obvious that our TE methods are more efficient than the traditional LDA-based measurement in time and memory cost. The reason is that our TE methods utilize a domain ontology to calculate the semantic similarity rather than a large LDA model. In general, our TE Methods that achieve semantic similarities based on extraction and template get desirable overall performance.

## 7. Conclusions

This paper proposes for the first time constructing a topic event to represent the semantics of a document and measures the semantic similarity between the academic literatures by computing the similarity of their corresponding topic events. We define the general architecture of topic event and give the construction and similarity computing methods of topic events. The research style ontology and an evaluation corpus are constructed. To measure the concept semantic similarity, we design the Computational Linguistics Ontology. The evaluation experiments show that our topic event method gets significantly improvement on current semantic similarity measurements, and the ontology-based TE methods show overall advantages in Correlation, Accuracy and F1-score. The knowledge resources such as the domain ontology play a crucial role in the document semantic similarity.

Furthermore, our method can be used for modeling the semantics of different styles of research work, and the ontology-based automatic construction and similarity calculation of TE are concordant across different domains, which means our method could be easily utilized in various academic domains.

## Acknowledgments

This research was supported by the Foundation of the State Key Laboratory of Software Development Environment (Grant No. SKLSDE-2015ZX-04). We thank the reviewers for their valuable feedback, which help to improve the paper.

## References

[1] Courtney Corley and Mihalcea Rada. 2005.: Measuring the Semantic Similarity of Texts. In: Proceedings of the ACL Workshop on Empirical Modeling of Semantic Equivalence and Entailment, pages 13-18.
[2] Rus, V., Lintean M., Graesser, A.C., & McNamara,D.S. 2009.: Assessing Student Paraphrases Using Lexical Semantics and Word Weighting. In: Proceedings of the 14th International Conference on Artificial Intelligence in Education, Brighton, UK.
[3] Banea C, Choi Y, Deng L, et al.: CPN-CORE: A Text Semantic Similarity System Infused with Opinion Knowledge. In: Proceeding of Second Joint Conference on Lexical and Computational Semantics. Atlanta, Georgia, USA, 2013: 221.
[4] Bill Dolan, Chris Quirk, and Chris Brockett. 2004.: Unsupervised Construction of Large Paraphrase Corpora: Exploiting Massively Parallel News Sources. In: Proceedings of ACL, pages 350.
[5] Agirre E, Banea C.: SemEval-2015 task 2: Semantic textual similarity, English, Spanish and pilot on interpretability[C] SemEval 2015, June.
[6] Daniel Ramage, Anna N. Rafferty, and Christopher D. Manning.: Random walks for text semantic similarity. In Proceedings of the 2009 workshop on graph-based methods for natural language processing. Association for Computational Linguistics, 2009: 23-31.
[7] Vasile Rus, Mihai Lintean, Cristian Moldovan, William Baggett, Nobal Niraula, and Brent Morgan.: The similar corpus: A resource to foster the qualitative understanding of semantic similarity of text. In: Proceedings of LREC, 2012:23-25.
[8] Frane Šarić, Goran Glavaš, Mladen Karan, Jan Šnajder and Bojana Dalbelo Bašić. 2012.: Takelab: Systems for measuring semantic text similarity. In: Proceedings of the Sixth International Workshop on Semantic Evaluation. Association for Computational Linguistics, papers 441-448.
[9] R. Amsler.: Application of citation-based automatic classification. Technical report, The University of Texas at Austin Linguistics Research Center, 1972.
[10] M. Kessler.: Bibliographic Coupling Between Scientific Papers, Journal of the American Documentation, Vol. 14, No. 1, pp.10-25, 1963.
[11] H. Small, "Co-citation in the Scientific Literature: A New Measure of the Relationship between Two Documents," Journal of the American Society for Information Science, Vol. 24, No.4, pp. 265-269, 1973.
[12] G Salton, A Wong, CS Yang.: A vector space model for automatic indexing[J], Communications of the ACM ,1975, V(11):613-620.
[13] David M. Blei, Andrew Y. Ng, and Michael I. Jordan. 2003.: Latent dirichlet allocation, J. of Mach Learn. Res., 3: 993-1022.
[14] Madylova A, Öğüdücü Ş G.: A taxonomy based semantic similarity of documents using the cosine measure[C]//Computer and Information Sciences. 24th International Symposium on. IEEE, 2009: 129-134.
[15] Nagwani N K, Verma S.: A frequent term and semantic similarity based single document text summarization algorithm [J]. International Journal of Computer Applications (0975–8887) Volume, 2011: 36-40.

[16] Schuhmacher M, Ponzetto S P.: Knowledge-based graph document modeling[C], In: Proceedings of WSDM, 2014: 543-552.
[17] Rafi M, Shaikh M S.: An improved semantic similarity measure for document clustering based on topic maps[J]. arXiv preprint arXiv:1303.4087, (2013)
[18] Wang Ying, Zhang Ru-bo, and Lai Ji-bao. 2009.: Measuring concept similarity between fuzzy ontologies. Fuzzy Information and Engineering 2: 163-171.
[19] Muyu Zhang, Bing Qin, Mao Zheng et al.: Encoding Distributional Semantics into Triple-Based Background Knowledge Ranking for Document Enrichment. In: Proceedings of ACL.
[20] Dragomir R. Radev, Pradeep Muthukrishnan, and Vahed Qazvinian. 2009.: The ACL anthology network corpus. In Proceedings of the 2009 Workshop on Text and Citation Analysis for Scholarly Digital Libraries. Association for Computational Linguistics, pages 54-61.
[21] Leacock, C., and Chodorow, M. 1998.: Combining local context and WordNet sense similarity for word sense identification. In WordNet, An Electronic Lexical Database. The MIT Press.
[22] Zhibiao Wu and Martha Palme.1994.: Verb semantics and lexical selection. In: Proceedings of ACL, pages 133-138.
[23] Lin, D. 1998.: An information-theoretic definition of similarity. In: Proceedings of the International Conf. on Machine Learning.
[24] Resnik, P. 1995.: Using information content to evaluate semantic similarity. In: Proceedings of the 14th International Joint Conference on Artificial Intelligence.
[25] Goikoetxea J, Soroa A, Agirre E, et al.: Random walks and neural network language models on knowledge bases[C]. In: Proceedings of NAACL-HLT. 2015: 1434-1439.
[26] Faruqui M, Dyer C.: Non-distributional word vector representations [J]. arXiv preprint arXiv:1506.05230, (2015)
[27] Thomas K. Landauer, Peter W. Foltz, and Darrell Laham. 1998.: An Introduction to latent semantic analysis. Discourse Processes, 25(2-3):259-284.
[28] J. Pennington, R. Socher, and C. D. Manning.: Glove: Global vectors for word representation. In: Proceedings of EMNLP, 2014.
[29] Turney, P. 2001.: Mining the web for synonyms: PMI-IR versus LSA on TOEFL. In: Proceedings of the Twelfth European Conference on Machine Learning.
[30] Cilibrasi, R.L. & Vitanyi, P.M.B. 2007.: The Google Similarity Distance, IEEE Trans. Knowledge and Data Engineering, 19:3, 370-383.
[31] Tomas Mikolov, Kai Chen, Greg Corrado, and Jeffrey Dean.: Efficient estimation of word representations in vector space. arXiv preprint arXiv:1301.3781, 2013.
[32] Quoc V Le and Tomas Mikolov.: Distributed representations of sentences and documents. arXiv preprint arXiv:1405.4053, 2014.
[33] Martín G H, Schockaert S, Cornelis C, et al.: Using semi-structured data for assessing research paper similarity [J]. Information Sciences, 2013, 221: 245-261.
[34] Muyu Zhang, Bing Qin, Ting Liu, et al.: Triple based Background Knowledge Ranking for Document Enrichment. In: Proceedings of COLING, 2014.
[35] Kim, J. H.; Kim, D.; Kim, S.; and Oh, A.: Modeling topic hierarchies with the recursive chinese restaurant process. In Proceedings of the 21st CIKM, 2012:783–792.
[36] Vasile Rus, Mihai Lintean, Rajendra Banjade, Nobal Niraula and Dan Stefanescu. 2013.: SEMILAR: The Semantic Similarity Toolkit. In: Proceedings of ACL, pages 163-168.
[37] Musen, M.A.: The Protégé project: A look back and a look forward. AI Matters. Association of Computing Machinery Specific Interest Group in Artificial Intelligence, 1(4), June (2015).
[38] Dominika Tkaczyk, Paweł Szostek, Mateusz Fedoryszak et al.: CERMINE: automatic extraction of structured metadata from scientific literature. In: International Journal on Document Analysis and Recognition (IJDAR), pp. 317–335, 2015.